# $L_1$-regularized neural ranking for risk stratification and its application to prediction of time to distant metastasis in luminal node negative chemotherapy naïve breast cancer patients

Fayyaz Minhas[1][0000-0001-9129-1189], Michael S. Toss[2], Noor ul Wahab[1], Emad Rakha[2] and Nasir M. Rajpoot[1][0000-0002-4706-1308]

[1]Tissue Image Analytics (TIA) Centre, Department of Computer Science, University of Warwick, Coventry, UK
[2]Nottingham Breast Cancer Research Centre, University of Nottingham, Nottingham, UK

{ fayyaz.minhas, noorul.wahab, n.m.rajpoot}@warwick.ac.uk
{ michael.toss, emad.rakha}@nottingham.ac.uk

**Abstract.** "*Can we predict if an early stage cancer patient is at high risk of developing distant metastasis and what clinicopathological factors are associated with such a risk?*" In this paper, we propose a ranking based censoring-aware machine learning model for answering such questions. The proposed model is able to generate an interpretable formula for risk stratification using a minimal number of clinicopathological covariates through $L_1$-regulrization. Using this approach, we analyze the association of time to distant metastasis (TTDM) with various clinical parameters for early stage, luminal (ER+/HER2-) breast cancer patients who received endocrine therapy but no chemotherapy (n = 728). The TTDM risk stratification formula obtained using the proposed approach is primarily based on mitotic score, histological tumor type and lymphovascular invasion. These findings corroborate with the known role of these covariates in increased risk for distant metastasis. Our analysis shows that the proposed risk stratification formula can discriminate between cases with high and low risk of distant metastasis ($p$-value < 0.005) and can also rank cases based on their time to distant metastasis with a concordance-index of 0.73.

## 1   Introduction

Using survival data for identification of important biomarkers or clinicopathological factors associated with survival and/or other clinically important events such as time to distant metastasis lies at the heart of clinical research [1]–[3] . Outcome based survival analysis is key for patient stratification into various subgroups for selection of personalized treatment options and analyzing the impact of therapeutic interventions [4], [5]. For this purpose, the most widely used approach is the Cox-Proportional Hazards model (Cox-PH) which relies on the assumption that each covariate has a multiplicative effect in the hazards function that is constant over time [6]. Regularization approaches for handling high dimensional covariates with the Cox-PH model have also been developed [7]. Alternative techniques such as regression or ranking based survival support vector machines (SSVMs) use L2 regularization of the weights and a squared error function which makes these models vulnerable to scaling issues and outliers in the data [8], [9]. Furthermore, mining survival data for extraction of a risk stratification formula based on a minimal set of covariates is complicated due to this regularization. Due to the structure of the loss function in regression based survival support vector machines, these models can potentially lead to over-estimation of survival times in censored cases. In the recent years, a large number of deep learning based survival prediction models have been developed [10]–[18]. However, difficulties associated with interpretability of these models makes identification of statistically significant markers for survival analysis and construction of risk stratification scores challenging [19]. In this work, we propose a simple ranking based formulation that uses $L_1$-regularization to obtain a minimal set of covariates to produce a risk stratification formula through bootstrap analysis. We demonstrate the effectiveness of the proposed approach over a dataset of Stage-1, luminal (ER+/HER2-) breast cancer patients who had been given endocrine therapy but no chemotherapy. Effective risk stratification of these patients can lead to personalized therapeutic interventions such as chemotherapy for these patients [20]–[22]. This cohort of patients is of significant clinical interest due to the differences in the impact of hormone therapy, chemotherapy and immunotherapy for these patients [23].

## 2 Materials and Methods

### 2.1 Dataset and Data Representation

The dataset used in this study was obtained from University of Nottingham with a total of 1300 patients which included $n = 728$ Stage-1 (Lymph Node Negative or LN0), ER+/HER2- patients that received endocrine therapy but no chemotherapy. The dataset contains, for each case, a set of clinicopathological covariates which include Multifocality, Invasive Tumor Size (in cms), the Nottingham Histological Grade and its components (Tubule Formation (T), Pleomorphism Score (P) and Mitosis Score (M)), Lymphovascular invasion (LVI), Associated DCIS and LCIS, Number of Positive Lymph nodes, Nottingham Prognostic Index (NPI), Stage, Estrogen Receptor (ER), Progesterone Receptor (PgR) and human epidermal growth factor receptor 2 (HER2) Status, Menopausal status and Patient age. In addition to these, the histological tumor type (HTT) of each case was assigned and grouped by a pathologist to be one of the 7 distinct prognostic types – $HTT_1$: No special type / NST (57.3% of all cases), $HTT_2$: Invasive Lobular Carcinoma (7.7%), $HTT_3$: Tubular and Tubular Mixed Carcinoma (17.1%), $HTT_4$: Mixed NST and Special Type carcinoma (10.2%), $HTT_5$: Other Special (Mucinous, Papillary, Micropapillary, Cribiform and Adenoid-Cystic Carcinoma, 2.0%), $HTT_6$: Mixed Lobular Carcinoma (4.5%) and $HTT_7$: Metaplastic Carcinoma (1.1%). We used one-hot-encoding to represent each histological tumor type as an indicator variable, resulting in a total of 23 clinicopathological vari-ables. We are also given the time to distant metastasis (TTDM) (in months) for these patients with a 15+ year follow-up (censored at 180 months) with a median follow-up time of 150 months.

### 2.2 Model Formulation and Implementation

For survival prediction, we assume a discovery subset $D = \{(x'_i, T_i, \delta_i), i = 1 \ldots |D|\}$ such that, for each case in it, we are given a vector of (un-scaled) covariates $x'_i$, an event indicator variable ($\delta_i$) which indicates whether the event of interest (such as distant metastasis or death) has taken place ($\delta_i = 1$) or not ($\delta_i = 0$) and the time $T_i$ to event (if $\delta_i = 1$) or censoring (if $\delta_i = 0$). All variables in the covariate set are scaled to the range [0,1] to eliminate effects of differences in their ranges to yield a vector $x_i$ for a given patient. This scaling also ensures that there are no prior assumptions on the importance of different covariates in terms of their impact on survival. Using the discovery set, we aim to obtain a prediction score $f(x; w) = \sigma(w^T x)$ for a given vector of covariates ($x$) using a weight vector $w$ where $\sigma(\cdot)$ is the bipolar sigmoid activation function. In order to train the proposed model, we first develop a dataset $P(D) = \{(i, j) | \delta_j = 1, T_i > T_j, \forall i, j = 1 \ldots |D|\}$ which contains pairs of comparable cases, i.e., all possible pairs of patients in the discovery set such that the event of patient with the shorter survival or censoring time has taken place. The weight vector is then obtained by solving the following optimization problem which generates a penalty if the predictor produces a higher score for a case with shorter survival time:

$$\min_{w} Q(w; D) = \lambda \|w\|_1 + \frac{1}{|P(D)|} \sum_{(i,j) \in P(D)} max\left(0, 1 - \left(f(x_i; w) - f(x_j; w)\right)\right)$$

where $\lambda \geq 0$ is a hyperparameter of the model (set to 0.01 for results in this paper) which controls the compromise between the degree of $L_1$-regularization and the average ranking loss over all comparable pairs. Note that this formulation differs from the classical ranking based survival support vector machine (SSVM) as it uses $L_1$-regularization which allows a smaller number set of covariates to have non zero weight values in comparison to L2-regularization used in SSVMs. Also, classical SSVMs uses a squared loss function which makes them vulnerable to outliers and survival time scaling issues. In contrast, the proposed model uses simple pairwise ranking hinge loss. The proposed model can be extended for use with non-linear kernel machines or deep learning models in which the prediction function $f(x; \theta)$ can be non-linear with lumped learnable parameters $\theta$. The above optimization problem can be optimized using gradient descent $\left(w \leftarrow w - \alpha \nabla_w Q(w; D)\right)$. We have implemented this model in PyTorch with adaptive momentum based optimization (Adam) which allows easy integration into deep learning pipelines as well.

### 2.3 Bootstrap Estimation and Risk Stratification

In order to get reliable risk stratification, we used a bootstrap estimation approach. Specifically, in each bootstrap run ($b = 1, 2, \ldots, B$), the given dataset is randomly divided into two subsets with event based stratification: a discovery subset (consisting of 60% cases) and a validation subset (consisting of 40% cases) such that the proportion of cases with events is the same in the two sets. In each bootstrap run, the discovery dataset is used to estimate model weights and rank cases in the validation set based on their survival times. The median of the predicted scores (of the training/discovery set) is used as a cut-off threshold to stratify cases in the validation set into risk categories. The log-rank test is used to estimate the *p*-value in each bootstrap run [24]. In order to combine multiple *p*-values from all runs, we have used the estimate $p_B = 2p_{50}$ where $p_{50}$ is the median of the p-values of the log-rank test across all bootstrap runs. This provides a conservative estimate of the overall p-value and, consequently, minimizes the chances of false discovery [25]. The concordance index of each run is also calculated as a metric for the quality of ranking based on time to distant metastasis [26]. An ideal ranking model will have a concordance index of 1.0 whereas a completely random one will have a concordance index of 0.5.

In order to obtain a score for risk stratification, the weight parameters from each bootstrap run $\{w^{(b)}, b = 1 \ldots B\}$ are rescaled with respect to their L$_1$-norm $\left(\frac{w^{(b)}}{\|w^{(b)}\|_1}\right)$ and the median of each element in the weight vector is used to rank different covariates. This yields a bootstrap estimate of the weight vector and all but the top-K weights (based on their magnitude) are set to zero. Inverse scaling is then applied to yield the final weight vector $w_K^*$ to adjust for the effect of the original range scaling and to convert the prediction scores $f(x_i; w)$, which are high (low) for high (low) survival times, to risk scores $r(x'; w_K^*)$ which are high for high risk (i.e., low survival) by multiplying the weights by -1.0 and adding an appropriate bias. The final weights can be used directly with unscaled covariates $x'$. Once the bootstrap estimate of the risk function is obtained, it is expressed as a simple equation leading to risk stratification and elucidation of the role of different covariates.

## 3   Results and Discussion

### 3.1   Bootstrap Analysis

The results of the bootstrap analysis (with *B*=1000 runs) are shown in Figure 1 which shows the distribution of the *p*-values along with the combined *p*-value ($p_B \ll 0.05$) and the concordance indices (mean c-index = 0.70). Figure 2 shows the distribution of weight values of all covariates in bootstrap runs in the form of violin plots. It is important to note that the magnitude of the median weight of the Mitosis score (M) is the highest in comparison to all other covariates. The negative value of the weight indicates that a high value of the mitosis score is negatively associated with time to distant metastasis. Thus, increased mitosis score is associated with higher risks of distant metastasis. Other clinicopathological parameters with high magnitude of weights are Lymphovascular invasion (LVI) and histological tumor type 6 (Mixed Lobular Carcinoma). Due to the use of L$_1$ regularization in the proposed method, most of the weight coefficients are close to zero. After rescaling, this analysis allows us to write a simplified scoring formula for risk of developing distant metastasis as: $R_{DM} =$

$0.26M + 0.48LVI + 0.38T_{MLC}$. This indicates that the risk of an individual is higher or equivalently the time to distant metastasis is shorter if their mitotic score ($M \in \{1,2,3\}$) is high or there is lymphovascular invasion $LVI \in \{0,1\}$ and/or if their histological tumor type is Mixed Lobular Carcinoma (Histological Tumor Type HTT 6, $T_{MLC} \in \{0,1\}$). Since the percentage of patients with MLC in our dataset is small (4.5%), the major contributors to the risk can be attributed to mitosis score and LVI.

### 3.2 Risk Stratification Index for Distant Metastasis and its Biological Interpretability

The resulting risk stratification function when applied to the whole cohort leads to very good risk stratification ($p \ll 0.05$) for time to distant metastasis as shown in Figure 3. The concordance index based on the risk function is 0.73. There is significant difference in the confidence intervals (shaded areas) of the survival probability curves for the high and low risk categories. This is also reflected in the number of events in each group at all time steps. Furthermore, using all covariates in the dataset has minimal impact on improving the concordance index in comparison to using only top 3 covariates (from 0.73 to 0.75).

The three variables involved in the risk stratification formula are known to have a significant impact on survival with a number of publications reporting the increased risk of distant metastasis associated with increased mitotic score [20], LVI [21] and mixed lobular carcinoma [27], [28] in this cohort. The reported formula provides a simplified expression for risk stratification based on these parameters.

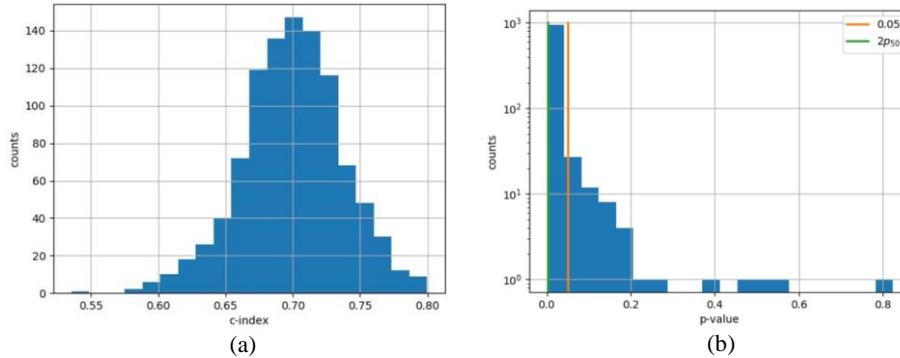

(a)  (b)

**Fig. 1.** Distributions of (a) concordance index and (b) p-values across bootstrap runs (note logarithmic y-axis in (b) to highlight low counts). The significance *p*-value cut-off of 0.05 (orange line) and the combined *p*-value ($2p_{50}$) of 0.002 (green line) are shown in (b).

**Table 1.** Result of bootstrap analysis of different methods with the combined p-value ($2p_{50}$) and average concordance index (standard deviation) of 1,000 bootstrap runs.

| Method | Combined *p*-value | Concordance Index (std) |
|---|---|---|
| $L_1$ Cox-PH | 0.007 | 0.68 (0.03) |
| Ranking SSVM | 0.010 | 0.68 (0.03) |
| Proposed | **0.002** | **0.70 (0.03)** |

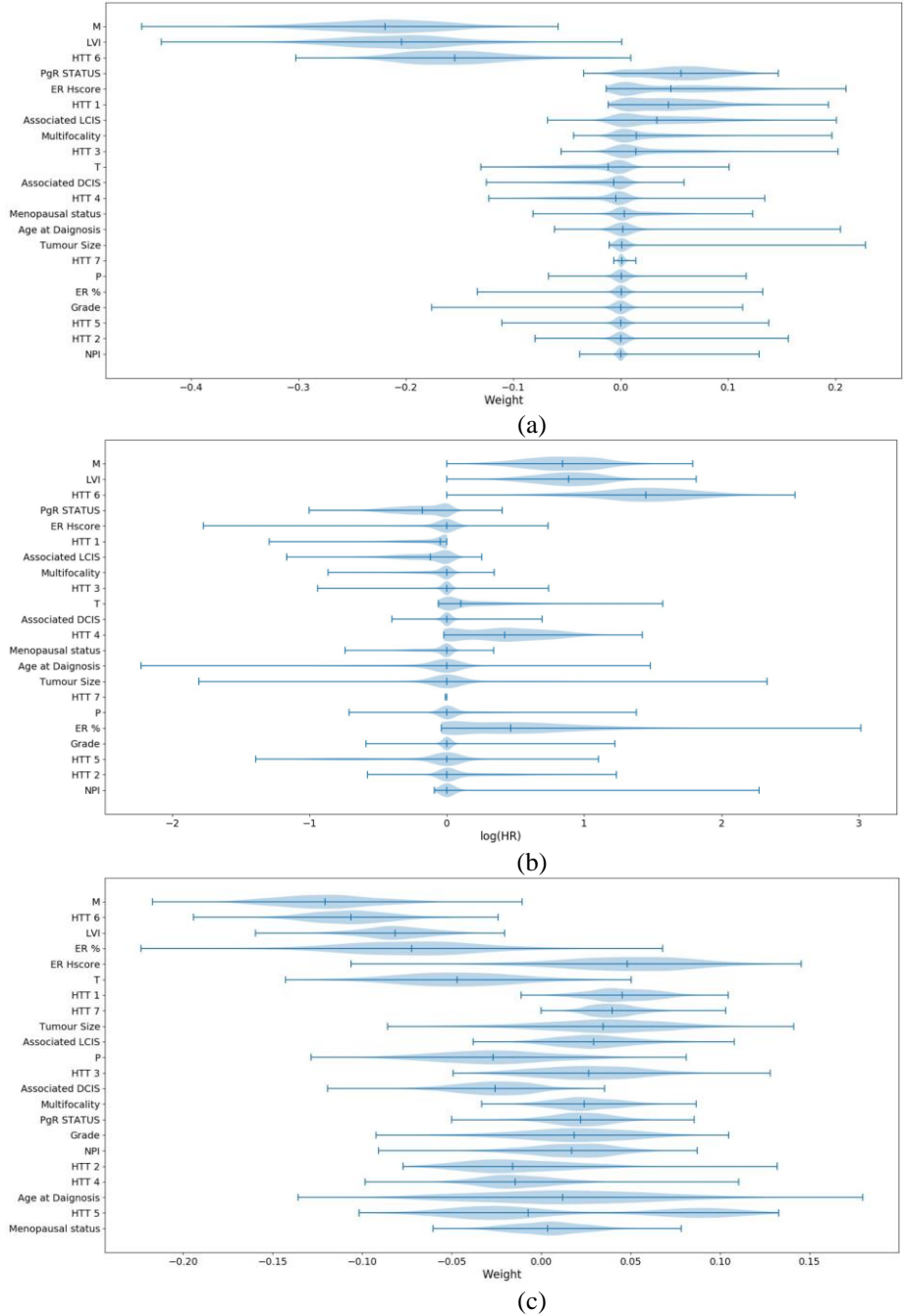

**Fig. 2.** Distributions of (a) weight values of the proposed model, (b) logarithm of the hazard ratios (HR) from $L_1$ regularized Cox-PH model and (c) weights of the ranking based survival SVM model for all covariates across bootstrap runs. All parameters ordered by importance.

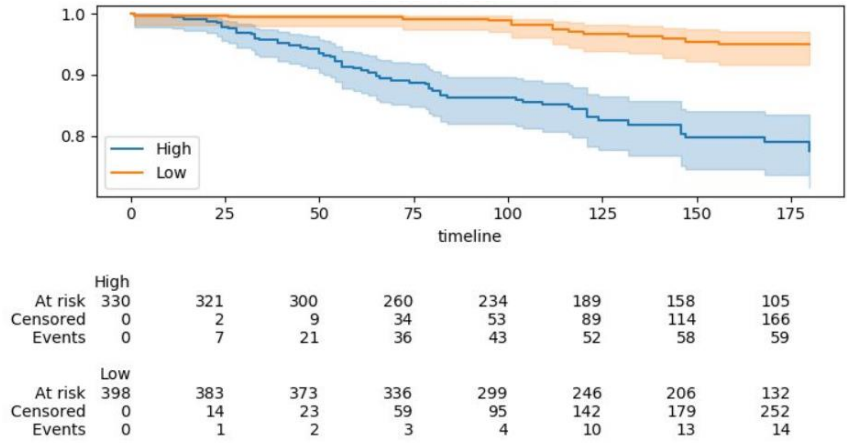

**Fig. 3.** Stratification of patients (n = 728) into low and high risk categories using the proposed risk stratification function and the resulting probability of distant metastasis free survival (y-axis) over time (months) (x-axis) (log-rank $p$-value = $9.1 \times 10^{-11}$).

### 3.3 Baseline and Comparison with other approaches

As concordance index can lead to over-estimation of prediction performance in cases with high censoring, we have first obtained a reference baseline by randomly shuffling event times of cases in the validation set and calculating the expected concordance index across all bootstrap runs. This results in a concordance index of 0.66 and can be interpreted as the expected minimal concordance of a model that predicts completely uninformative survival scores. The concordance index of the proposed model (0.73) is significantly higher in comparison to this baseline.

We have also performed a comparison of the proposed method with a ranking based Survival-SVM (SSVM) and $L_1$-regularized Cox-Proportional Hazard (Cox-PH) model. $L_1$ regularization was used in the Cox-PH model to avoid low-rank matrix inversion issues due to indicator variables in the data and to provide a direct and fair comparison with $L_1$-regularization in the proposed model. Un-regularized Cox-PH is not applicable for this dataset due to high collinearity resulting from the use of indicator variables. In order to avoid differences due to data scaling, each method was given the rescaled dataset. The distributions of weight values for the SSVM and the hazard ratios for different covariates from the Cox-PH model are shown in Figure 2. In line with the proposed model, both these models have also identified mitosis score, LVI and mixed lobular carcinoma (HTT 6) as the most important factors in prediction of time to distant metastasis. However, in comparison to the proposed model, both Cox-PH and SSVM show possible contributions from a larger number of other variables (relative to their scaling of hazard ratios and weights, respectively) which can complicate the development of a risk stratification formula. Furthermore, the SSVM model requires scaling of event times due to its use of the square loss function which makes it very sensitive to outliers. The combined $p$-value and c-index of the bootstrap runs are reported in table-1 which shows that the proposed method can potentially reduce false discovery rate (lower combined *p*-values) and improve ranking quality (higher concordance indices).

The primary advantage of the use of the proposed method in comparison to these approaches is its ability to provide a risk stratification formula without relying on any proportionality of hazard assumption and explicitly reduce the number of covariates in the resulting formula. Furthermore, the proposed approach can be easily integrated with deep neural network architectures for end-to-end survival prediction.

### 3.4   Univariate Analysis of top covariates

In order to verify that the top covariates identified by the proposed approach are indeed statistically significant for risk stratification of distant metastasis, we also performed univariate analysis based on these covariates. The *p*-values of the log-rank test for stratification based on LVI (LVI Detected vs Not Detected), Mitosis score (M = 1 vs. M > 1) and Histological Tumor Type (MLC (HTT-6) vs. All others (non-HTT-6)) are $9.6 \times 10^{-5}$, $6.5 \times 10^{-6}$ and $1.5 \times 10^{-4}$, respectively. It shows that all these top covariates show statistically significant evidence of correct risk stratification with low chances of false discovery over this cohort. It is important to note that these variables show statistical significance for censoring times of 60, 120, 180 and 250 months as well. The corresponding Kaplan-Meier curves of these variables are shown in Figure 4 and indicate clear distinction between different groups.

## 4   Conclusions and Future Work

In this work, we have proposed an interpretable ranking based approach with $L_1$ regularization which can be used for identification of a minimal set of important clinicopathological parameters in survival data. Using a dataset of luminal breast carcinomas, we have demonstrated that the proposed approach is able to recover a minimal set of covariates and develop a risk score. The bootstrap estimate of the risk to distant metastasis developed in this study highlights the role of mitotic score, lymphovascular invasion and histological tumor type and can potentially be of clinical and therapeutic interest after a large-scale validation. These results can also be used as a baseline for the development of AI approaches which can overcome inherent subjectivity in the assessment of these pathological parameters. The proposed approach due to its differentiable nature can be extended further for integration into deep learning models in an end-to-end manner as well.

9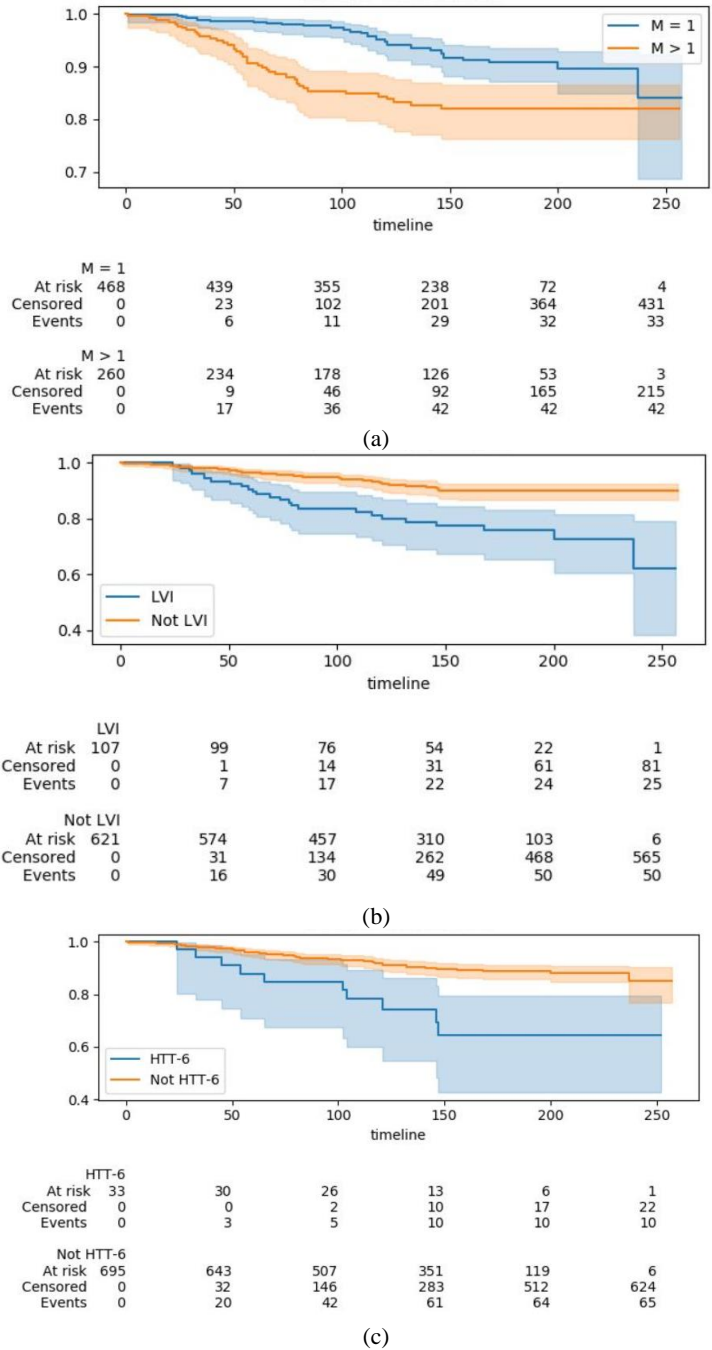

**Fig. 4.** Results of univariate analysis with (a) mitosis score, (b, LVI (middle) and (c) Histological Tumor Type 6 (MLC). Each curve shows the probability of distant metastasis free survival (y-axis) across time (months) (x-axis).